\documentclass[letterpaper, 10 pt, conference]{ieeeconf}  

\IEEEoverridecommandlockouts                              

\usepackage{tabularx}
\usepackage{graphics} 
\usepackage{amsmath} 
\usepackage{amssymb}  
\usepackage{bm}         
\makeatletter
\renewcommand{\maketag@@@}[1]{\hbox{\m@th\normalsize\normalfont#1}}%
\makeatother
\title{\LARGE \bf
Event-based Stereo Visual-Inertial Odometry
with Voxel Map
}

\author{
  Zhaoxing Zhang$^{1}$, 
  Xiaoxiang Wang$^{1}$, 
  Chengliang Zhang$^{1}$, 
  Yangyang Guo$^{2}$, 
  Zikang Yuan$^{1}$, 
  Xin Yang$^{1}$\thanks{Corresponding authors: Zikang Yuan and Xin Yang.}
}

\usepackage{booktabs}
\usepackage{threeparttable} 
\usepackage{multirow}
\usepackage{array}

\usepackage{algorithmic}
\usepackage{algorithm}
\usepackage{bm}

\usepackage{hyperref}
\usepackage{makecell}
\usepackage{cite}

\usepackage{graphicx}

\begin{document}

\maketitle
\thispagestyle{empty}
\pagestyle{empty}

\begin{abstract}

The event camera, renowned for its high dynamic range and exceptional temporal resolution, is recognized as an important sensor for visual odometry. However, the inherent noise in event streams complicates the selection of high-quality map points, which critically determine the precision of state estimation. To address this challenge, we propose Voxel-ESVIO, an event-based stereo visual-inertial odometry system that utilizes voxel map management, which efficiently filter out high-quality 3D points. Specifically, our methodology utilizes voxel-based point selection and voxel-aware point management to collectively optimize the selection and updating of map points on a per-voxel basis. These synergistic strategies enable the efficient retrieval of noise-resilient map points with the highest observation likelihood in current frames, thereby ensureing the state estimation accuracy. Extensive evaluations on three public benchmarks demonstrate that our Voxel-ESVIO outperforms state-of-the-art methods in both accuracy and computational efficiency.
\end{abstract}

\section{INTRODUCTION}
\label{Introduction}

\begin{figure}
\centering
\includegraphics[width=0.47\textwidth]
{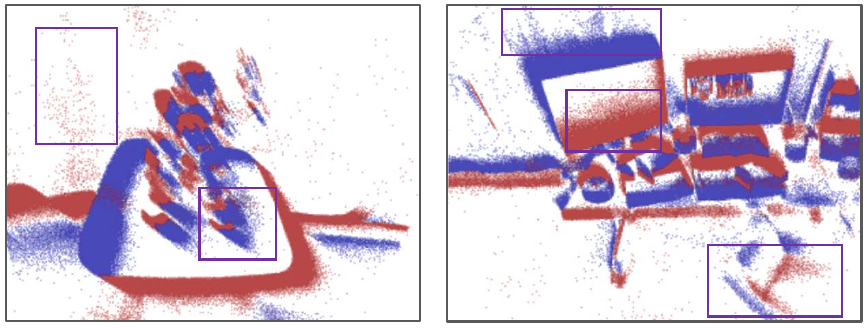}
\caption{Illustration of event noise: Red and blue events indicate positive and negative polarity signals triggered by the event camera, respectively.}
\label{fig:event}
\end{figure}

As a novel bio-inspired vision sensor, the event camera operates asynchronously to detect pixel-level brightness changes and output sparse event streams, which provide ultra-high temporal resolution and exceptionally wide dynamic range \cite{event_camera0,event_camera1,event_camera2}. In contrast to visual-inertial odometry (VIO) \cite{qin2018vins, campos2021orb, voxel-svio, standard_pipeline2, standard_pipeline3} and LiDAR-inertial odometry (LIO) \cite{yuan2022sr, yuan2024sr, yuan2023sdv, yuan2023semi}, event-inertial odometry (EIO) \cite{evo,feature_based,esvio,esvio2,esvo,esvio_aa,esvo2} demonstrates unique advantages in challenging conditions with rapid motion or dramatic lighting changes. However, the event-driven mechanism is highly sensitive to noise—even minor sensor vibrations or slight illumination changes can trigger spurious events (as shown in Fig. \ref{fig:event}). When early event-based approaches \cite{evo, early_work1, feature_based, track_enhance1} directly fed these noise-contaminated event frames into the VO pipeline, they suffered from incorrect feature matches, erroneous map points, and degraded performance.

Recent research has sought to mitigate these effects by enhancing the event stream tracking module—either through improved event representation methods\cite{esvo2, esvio_aa, invarient_representation1, invarient_representation2,esvio} or more robust event frame matching strategies \cite{esvio2,track_enhance1,pl-evio,track_enhance2,track_enhance3}. While these approaches have made notable progress, they still exhibit suboptimal performance in challenging scenarios. In this work, we focus on the selection of high-quality map points for optimizer integration to mitigate the propagation of event noise through subsequent odometry modules. We contend that high-quality points should satisfy two essential criteria:
1) Observability: Points should be reliably observable by current frames to support the formation of valid geometric constraints.
2) Noise Resilience: Points must originate from genuine object-edge movement rather than event noise artifacts.
For observability, current approaches typically rely on threshold-based filtering of individual residuals to select map points \cite{esvo, esvio_aa, esvo2, ultimate, esvio}, which fundamentally fails to ensure point observability in current frames. For noise resilience, the global management of map points in most existing methods leads to the projection of both valid and invalid feature matches into 3D space during map point registration, without removing false correspondences caused by event noise.

\begin{figure}
\centering
\includegraphics[width=0.47\textwidth]
{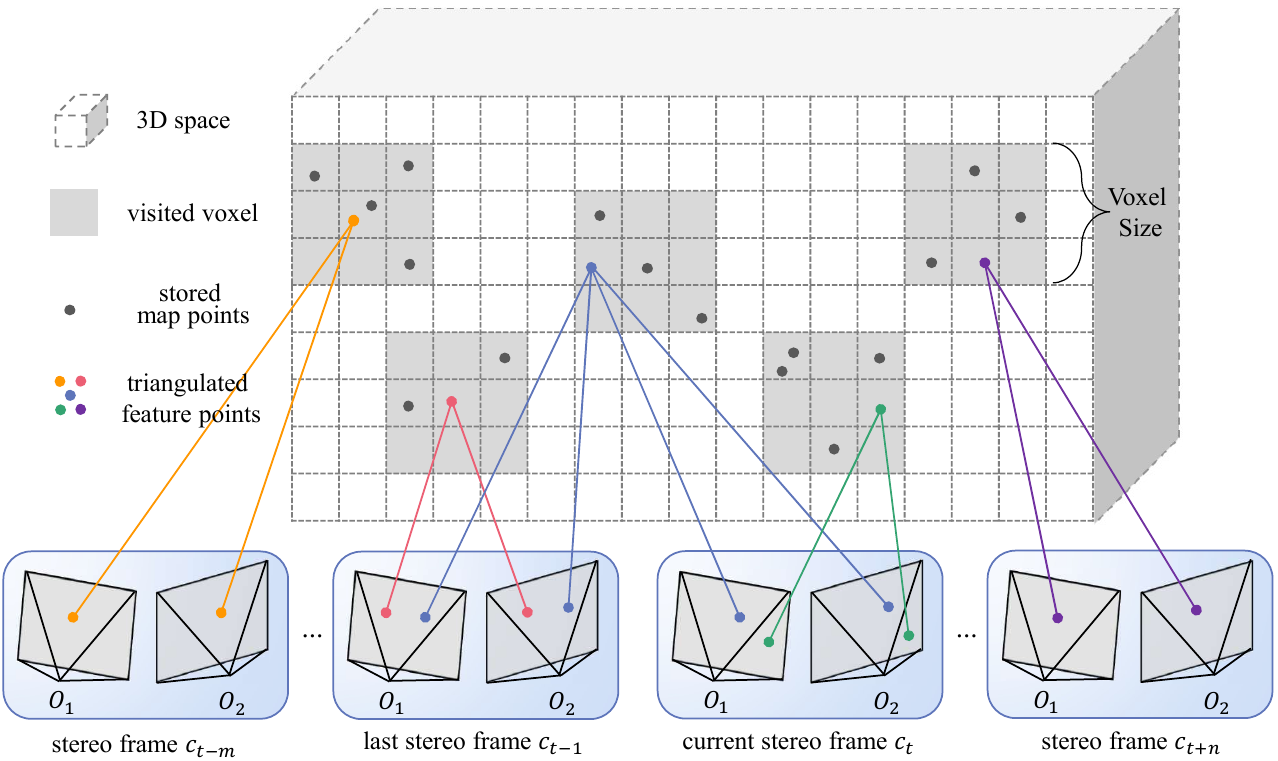}
\caption{Illustration of voxel map management for map point organization. Through triangulation of feature correspondences in the current stereo frame, we can directly index to the corresponding voxels in space.}
\label{fig:first}
\end{figure}

In this work, we propose Voxel-ESVIO, an event-based stereo visual-inertial odometry system that introduces voxel map management to select high-quality map points to mitigate the propagation of event noise. As illustrated in Fig. \ref{fig:first}, voxel map management enables direct indexing of voxels through triangulated feature correspondences from the current stereo event frame. This voxel-based point selection strategy prioritizes the most observationally relevant 3D map points, thereby maximizing observation likelihood from the current viewpoint. To further enhance mapping quality, we introduce a voxel-aware point management strategy featuring a three-stage cascaded filtering pipeline: 1) temporal consistency validation, 2) spatial proximity culling, and 3) voxel capacity regulation. The first two criterion optimize point distribution across temporal and spatial domains respectively, while the third enforces uniform 3D point density to maintain computational efficiency. Experimental results on three public datatests demonstrate the outstanding accuracy and computational efficiency of our system.

Our contributions can be summarized as follows:
\begin{itemize}
\item We integrate a voxel map management for EIO framework with two core strategies, i.e., voxel-based point selection and voxel-aware point management, which mitigate the propagation of event noise.
\item Our framework demonstrates superior accuracy and computational efficiency over state-of-the-art methods across three public datasets.

\end{itemize}

The remainder of this paper is organized as follows: Sec. \ref{section:related_work} reviews related work, Sec. \ref{section:methodology} introduces our proposed Voxel-ESVIO system, Sec. \ref{section:experiment} presents experimental evaluations, and Sec. \ref{section:conclusion} concludes the paper.

\section{RELATED WORKS}
\label{section:related_work}
Similar to conventional frame-based visual odometry systems\cite{standard_pipeline1}, event-based visual odometry addresses the coupled problems of camera tracking and 3D mapping through recursive estimation. Depending on the mechanism for data association, existing methodologies broadly fall into two categories: 1) direct methods and 2) feature-based indirect methods.

\subsection{Direct Methods}
Kim \textit{et al.} \cite{construct_intensity} pioneered the first event-based direct method, utilizing three interleaved probabilistic filters for pose, depth and intensity estimation. However, \cite{construct_intensity} required intensive GPU processing for intensity reconstruction and depth regularization, which limited its lightweight application. EVO\cite{grupp2017evo} introduced a more efficient approach running entirely on CPU. The system detected 3D structures by identifying maximum ray intersections in the disparity space image (DSI)\cite{emvs} and solved camera pose through 3D-2D registration by aligning the 3D map with the synthesized event map. Building on EVO, ESVO\cite{esvo} developed the first stereo event-based VO framework. By integrating both spatial and temporal consistency of event image, ESVO successfully improved accuracy in localization and mapping.
Several methods then enhanced ESVO by incorporating IMU information. ESVIO\cite{esvio2} added IMU pre-integration constraints in the backend to improve pose estimation accuracy. ESVIO\_AA\cite{esvio_aa} used pre-integrated IMU gyroscope measurements as prior information to reduce the degradation of pose estimation and introduced an innovative adaptive accumulation event representation method.
ESVO2\cite{esvo2} then streamlined the system's computational requirements by implementing a fast block matching algorithm and simplifying backend optimization.
Although direct methods have advanced rapidly in recent years, existing approaches still show unstable performance and high computational demands when running in complex environments. System failures commonly occur due to insufficient mapping and large-baseline data association. 

\subsection{Indirect Methods}

To accommodate the unique characteristics of event data, researchers have proposed various hand-crafted feature extraction techniques \cite{fast_event,fast_event2} and event feature tracking algorithms \cite{track_enhance1,track_enhance2,track_enhance3}. \cite{track_enhance1} is the frist work using feature-tracks to achieve event-based VO, which detects  features from event images and then uses tracked features asynchronously. Later, \cite{mono-eio} introduced a real-time monocular event-based VIO system leveraging graph optimization, where raw asynchronous events were directly utilized for feature detection. PL-EVIO \cite{pl-evio} advanced this direction by fusing data from standard and event cameras, tightly integrating event-based point and line features, image-based point features, and inertial measurements. However, due to the motion-dependent and noise-sensitive nature of event streams, these features often lack robustness when tracking in real-world scenarios. To address this, incorporating IMU measurements has proven beneficial. \cite{real-time} improved feature quality through motion compensation, aligning events using IMU-estimated motion. 
Building upon this, Ultimate-SLAM \cite{ultimate} presented a unified framework that tightly integrates event cameras, standard cameras, and IMU data. This approach was further extended by ESIO \cite{esvio}, which incorporated stereo event and standard cameras along with IMU measurements for accurate and robust state estimation.
Compared to direct methods, indirect approaches generally offer greater robustness and stability, with reduced risk of failure in scenarios requiring large-baseline data association. However, the performance of indirect methods critically depends on feature correspondences, rendering them vulnerable to event noise-induced influence. To mitigate this limitation, we propose voxel-based map management, which reduces the dependency on explicit feature correspondences. This approach significantly improves the robustness of indirect methods in high-noise scenarios while maintaining computational efficiency.

\begin{figure*}[t]
\centering
\includegraphics[width=0.97\linewidth]{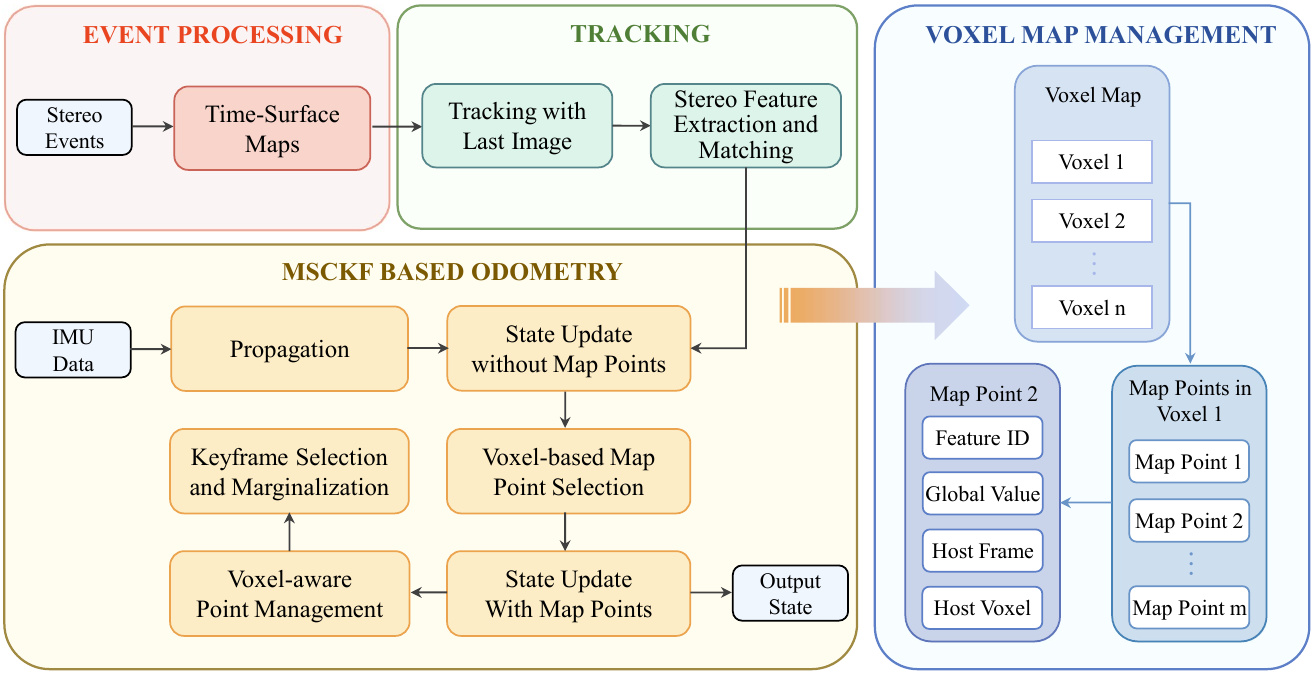}
\caption{System framework of Voxel-ESVIO, which comprises four main modules: the Event Processing Module, Tracking Module, MSCKF based Odometry Module, and Voxel Map Management Module.}
\label{fig:Overview}%
\end{figure*}

\section{METHODOLOGY}
\label{section:methodology}

\subsection{System Overview} 
Fig. \ref{fig:Overview} illustrates the overall framework of our system, which includes four main modules. The event processing module converts raw stereo event streams into image-like representations. The tracking module conducts feature correspondences across left and right camera of different frames. These feature correspondences are sent to the MSCKF-based odometry module, which estimates the state of recent frames and the global value of map points in real time. Finally, the voxel map management module maintains a structured 3D environmental representation by connecting voxels with both 3D landmarks and 2D feature correspondences across frames, facilitating efficient organization and real-time map point updates.

\subsection{Event Processing} 
Event cameras output a continuous stream of asynchronous events. Each event $e_k = (u_k,v_k,t_k,p_k)$ contains spatial and temporal coordinates that mark where an intensity change exceeds a specific threshold, along with a polarity value $p_k \in \{ +1, -1\}$ that indicates whether the intensity increased or decreased.
We convert these raw event streams into image-like representations called time-surface maps\cite{time-surface}. These time surface (TS) maps use an exponential decay kernel $\eta$ to preserve the temporal coherence of the event data. The TS map at time $t$ is defined by
\begin{equation}
\mathcal{T}(\mathbf{x}, t) \doteq \exp \left(-\frac{t-t_{\mathrm{last}}(\mathbf{x})}{\eta}\right) ,
\end{equation}
where $t_{\mathrm{last}}$ is the timestamp of the last event at each pixel coordinate $x = (u,v)^T$.
\subsection{Tracking}
\subsubsection{Tracking with Last Image}
When a new stereo time-surface (TS) map\cite{esvo} is generated, we use event-corner features from the previous left TS map to perform Kanade-Lucas-Tomasi (KLT) optical flow tracking\cite{klt} on the current left TS map. Event-corner features that fail to track in the left TS map are immediately discarded.

\subsubsection{Stereo Feature Extraction and Matching}
To maintain sufficient features for tracking, new features are added when tracked features are lost over time. The process begins by masking all currently tracked features in the current left image, then applying \cite{goodfeatures} to the unmasked regions. The system then performs KLT optical flow tracking on the current right image, using both the original tracked features and newly detected features to complete stereo matching.

\subsection{Voxel Map Management}
\label{sec:voxel_map_management}
As shown in the voxel map management module in Fig. \ref{fig:Overview}, we detail the specific implementation of our system for map point management. The 3D space is divided into a voxel map, which consists of numerous voxels. Each voxel stores a collection of map points. Every map point preserves four attributes: feature ID, global value, host frame, and host voxel.
The feature ID describes the feature correspondence of the point across multiple event frames during tracking. The global value represents the point’s 3D coordinates in the global coordinate system.
The host frame refers to the reference keyframe where the point was initially generated through triangulation. The host voxel indicates the voxel in which the map point resides. By using these concise elements, the voxel-based representation is seamlessly integrated into the system framework. This enables the transition from a purely point-based architecture to a voxel-partitioned framework, allowing more efficient map point management.

\subsection{MSCKF based Odometry}

The state vector (denoted as $\boldsymbol{\mathrm{x}}_k$) of our system comprises: the current inertial navigation state (denoted as $\boldsymbol{\mathrm{x}}_I$), a set of $n$ historical pose clones within the sliding window (denoted as $\boldsymbol{\mathrm{x}}_C$), a set of $m$ global 3D coordinates of selected map points (denoted as $\boldsymbol{\mathrm{x}}_M$), the extrinsic and intrinsic parameters of stereo cameras (denoted as $\boldsymbol{\mathrm{x}}_{Y_0}$ and $\boldsymbol{\mathrm{x}}_{Y_1}$), and the time-shift between the IMU and camera clocks (denoted as ${}^C t_I$).

\begin{equation}
\boldsymbol{\mathrm{x}}_k=\left[\begin{array}{llllll}
{\boldsymbol{\mathrm{x}}_I}^T & {\boldsymbol{\mathrm{x}}_C}^T & {\boldsymbol{\mathrm{x}}_M}^T & {\boldsymbol{\mathrm{x}}_{Y_0}}^T & 
{\boldsymbol{\mathrm{x}}_{Y_1}}^T & {}^C t_I

\end{array}\right]^T \\
\end{equation}

\begin{equation}
\boldsymbol{\mathrm{x}}_I=\left[\begin{array}{lllll}
{{}_G^{I_k} \boldsymbol{\mathrm{q}}}^T & {{}^G \boldsymbol{\mathrm{p}}_{I_k}}^T & {{}^G \boldsymbol{\mathrm{v}}_{I_k}}^T & {\boldsymbol{\mathrm{b}}_{\omega_k}}^T & {\boldsymbol{\mathrm{b}}_{a_k}}^T
\end{array}\right]^T \\
\end{equation}

\begin{small} 
\begin{equation}
\boldsymbol{\mathrm{x}}_C = \left[\begin{array}{c@{\quad}c@{\quad}c@{\quad}c@{\quad}c}
{{}_G^{I_{k-1}} \boldsymbol{\mathrm{q}}}^T & {{}^G\boldsymbol{\mathrm{p}}_{I_{k-1}}}^T & \vdots & {{}_G^{I_{k-n}}\boldsymbol{\mathrm{q}}}^T & {^G\boldsymbol{\mathrm{p}}_{I_{k-n}}}^T
\end{array}\right]^T
\end{equation}
\end{small}

\begin{equation}
\boldsymbol{\mathrm{x}}_M=\left[\begin{array}{llll}
{^G\boldsymbol{\mathrm{p}}_{f_1}}^T & {^G\boldsymbol{\mathrm{p}}_{f_2}}^T & \cdots & {^G\boldsymbol{\mathrm{p}}_{f_m}}^T
\end{array}\right]^T \\
\end{equation}

\begin{equation}
\boldsymbol{\mathrm{x}}_{Y_0}=\left[\begin{array}{llll}
{^I_{C_0}\boldsymbol{\mathrm{q}}}^T & {^{C_0}\boldsymbol{\mathrm{p}}_I}^T & {\boldsymbol{\mathrm{F}}_0}^T & \boldsymbol{\mathrm{D}}_0^T
\end{array}\right]^T \\
\end{equation}

\begin{equation}
\boldsymbol{\mathrm{x}}_{Y_1}=\left[\begin{array}{llll}
{^I_{C_1}\boldsymbol{\mathrm{q}}}^T & {^{C_1}\boldsymbol{\mathrm{p}}_I}^T & {\boldsymbol{\mathrm{F}}_1}^T & \boldsymbol{\mathrm{D}}_1^T
\end{array}\right]^T \\
\end{equation}


Here, $\boldsymbol{\mathrm{q}}$ denotes the rotational transformation, which can be converted into the corresponding rotation matrix $\boldsymbol{\mathrm{R}}$. The vector $\boldsymbol{\mathrm{p}}$ represents the translation between coordinate frames, while $\boldsymbol{\mathrm{v}}$ captures the linear velocity of the system. The vectors $\boldsymbol{\mathrm{b}_\omega}$ and $\boldsymbol{\mathrm{b}_a}$ correspond to the gyroscope and accelerometer biases, respectively.
The 3D position of each map point in the global coordinate frame is denoted by $\boldsymbol{\mathrm{p}}$. The camera's intrinsic parameters are represented by the focal length vector $\boldsymbol{\mathrm{F}}$, and the distortion parameter vector $\boldsymbol{\mathrm{D}}$ accounting for lens-induced geometric distortions during image formation.
We use superscripts to indicate coordinate frames: $(\cdot)^C$ for the camera, $(\cdot)^I$ for the IMU, and $(\cdot)^G$ for the global coordinate system. At initialization, the global frame $(\cdot)^G$ is aligned with the IMU frame $(\cdot)^I$, establishing a consistent reference frame for all subsequent transformations.

For vector variables, the "boxplus" and "boxminus" operations serve to map elements to and from a given manifold. They correspond to the straightforward addition and subtraction of their respective vectors:

\begin{equation*}
\mathbf{R}, \mathbf{R}_1, \mathbf{R}_2 \in SO(3); \mathbf{r} \in so(3);\boldsymbol{a}, \boldsymbol{b} \in \mathbb{R}^n; 
\end{equation*}
\vspace{-0.15em} 
\begin{equation}
\mathbf{R} \boxplus \mathbf{r}=\mathbf{R} \operatorname{Exp}(\mathbf{r});  \mathbf{R}_1 \boxminus \mathbf{R}_2=\log \left(\mathbf{R}_2^{\mathrm{T}} \mathbf{R}_1\right)
\end{equation}
\vspace{-0.15em} 
\begin{equation*}
\mathbf{a} \boxplus \mathbf{b}=\mathbf{a}+\mathbf{b}; \mathbf{a} \boxminus \mathbf{b}=\mathbf{a}-\mathbf{b}
\end{equation*}


\subsubsection{Propagation}
The current inertial state $\mathbf{x}_I$ is propagated forward using incoming IMU measurements of linear acceleration $^I\mathbf{a}_m$ and angular velocity $^I\mathbf{w}_m$. This propagation follows a generic nonlinear IMU kinematic model, which updates the state from time step $k-1$ to $k$ as follows:
 
\begin{equation}
\boldsymbol{x}_k^I=f\left(\boldsymbol{x}_{k-1}^I, ^I\boldsymbol{a}_m, ^I\boldsymbol{\omega}_m, \mathbf{n}\right)
\end{equation}
where $\mathbf{n}$ accounts for the IMU measurement noise, including both zero-mean white Gaussian noise and random walk bias noise. The predicted state estimate at the current time step is computed as:

\begin{equation}
\hat{\boldsymbol{x}}_{k \mid k-1}^I=f\left(\hat{\boldsymbol{x}}_{k-1 \mid k-1}^I, ^I\boldsymbol{a}_m, ^I\boldsymbol{\omega}_m, \boldsymbol{0}\right)
\end{equation}
where $\left( \hat{\cdot} \right)$ denotes the predicted estimate, and the subscript $k | k-1$ indicates the estimate at time step $k$ given measurements up to time $k-1$. The state covariance matrix is typically propagated by linearizing the nonlinear system model around the current predicted estimate:

\begin{equation}
\mathbf{P}_{k \mid k-1}=\boldsymbol{\Phi}_{k-1} \mathbf{P}_{k-1 \mid k-1} \boldsymbol{\Phi}_{k-1}^T+\mathbf{Q}_{k-1}
\end{equation}
where $\boldsymbol{\Phi}_{k-1}$ and 
$\mathbf{Q}_{k-1}$  represent the system Jacobian matrix and discrete noise covariance matrix, respectively\cite{jacobian}.

\subsubsection{State Update without Map Points}
\label{subsection:without_maps}
During the update phase, the optimization framework focuses solely on feature correspondences across multiple keyframes within a sliding window, explicitly avoiding reliance on previously mapped landmarks. To strike a balance between computational efficiency and estimation robustness, the system prioritizes the triangulation and refinement of the top $n$ feature points with the highest co-visibility across frames. As no persistent map points are used in this process, the state vector is simplified to include only the pose parameters of the selected keyframes, formulated as:
\begin{equation}
\boldsymbol{x}_k=\left[\begin{array}{lllll}
\boldsymbol{x}_I^T & \boldsymbol{x}_C^T & \boldsymbol{x}_{Y_0}{ }^T & \boldsymbol{x}_{Y_1}^T & t_I^C
\end{array}\right]^T
\end{equation}
Consider the following nonlinear measurement function:

\begin{equation}
\mathbf{z}_{\boldsymbol{w}, k}=h\left(\boldsymbol{x}_k^{s u b}\right)+\mathbf{n}_{\boldsymbol{w}, k}
\end{equation}
where $h(\cdot):\mathbb{R}^3\rightarrow\mathbb{R}^2$ denote the perspective projection mapping from triangulated 3D points to the 2D image plane. The term $\mathbf{z}_{w,k}$ represents the 2D feature observation of the corresponding 3D point in frame $k$. The measurement noise $\mathbf{n}_{w,k}$ is modeled as zero-mean Gaussian noise: $\mathbf{n}_{w,k} \sim \mathcal{N}(\mathbf{0}, \mathbf{R}_{w,k})$, where $\mathbf{R}_{w,k}$ is a diagonal covariance matrix with unit eigenvalues in our implementation. Following the linearized measurement model\cite{ekf}, the state update can be expressed as:

\begin{equation}
\hat{\boldsymbol{x}}_{k \mid k}^{\text {sub }}=\hat{\boldsymbol{x}}_{k \mid k-1}^{\text {sub }} \boxplus \mathbf{K}_k\left(\mathbf{z}_{\boldsymbol{m}, k}-h\left(\hat{\boldsymbol{x}}_{k \mid k-1}^{s u b}\right)\right) \\
\end{equation}
\begin{equation}
\mathbf{P}_{k \mid k}^{\prime}=\mathbf{P}_{k \mid k-1}-\mathbf{K}_k \mathbf{H}_k \mathbf{P}_{k \mid k-1} \\
\end{equation}
\begin{equation}
\mathbf{K}_k=\mathbf{P}_{k \mid k-1} \mathbf{H}_k^T\left(\mathbf{H}_k \mathbf{P}_{k \mid k-1} \mathbf{H}_k^T+\mathbf{R}_{\boldsymbol{m}, k}\right)^{-1}
\end{equation}
where $\mathbf{H}$ is the Jacobian respect to $\boldsymbol{x}_{k}^{\text {sub}}$ here.

\subsubsection{Voxel-based Point Selection}
\label{sec:select_map_points}
To ensure reliable geometric constraints during state update, our system selects map points based on a voxelized spatial representation. Using the updated keyframe poses from Sec. \ref{subsection:without_maps}, feature correspondences in the current stereo images are first triangulated to recover their 3D coordinates. These 3D points are then used to efficiently index the corresponding voxels in the voxel map.
Voxel access is restricted to those intersecting the frustum of the current stereo camera, as these spatial units correspond to scene regions that are projected onto the image plane—thereby maximizing the likelihood of establishing meaningful observations in subsequent frames.
To enhance scene coverage and minimize redundancy, the selection strategy incorporates not only each primary voxel but also its second-order topological neighbors. Map points are sampled from each selected voxel as well as its adjacent voxels, ensuring a balance between geometric relevance and spatial distribution. This combination of frustum-guided prioritization and neighborhood-based expansion enables the system to select map points that contribute effectively to the next state update.

\subsubsection{State Update with Map Points}
In this update step, all variables contained in $x_k$ are refined using the 3D map points obtained from the previous step. 
We construct geometric constraints within the sliding window using 2D feature correspondences and these 3D points.
The nonlinear measurement function is formulated as:
\begin{equation}
\mathbf{z}_{\boldsymbol{m}, k}=h\left(\boldsymbol{x}_k\right)+\mathbf{n}_{\boldsymbol{m}, k}
\end{equation}
Then the state update is performed as:
\begin{equation}
\hat{\boldsymbol{x}}_k=\hat{\boldsymbol{x}}_k \boxplus \mathbf{K}_k\left(\mathbf{z}_{m, k}-h\left(\hat{\boldsymbol{x}}_k\right)\right) \\
\end{equation}
\begin{equation}
\mathbf{P}_{k \mid k}=\mathbf{P}_{k \mid k}^{\prime}-\mathbf{K}_k \mathbf{H}_k \mathbf{P}_{k \mid k}^{\prime} \\
\end{equation}
\begin{equation}
\mathbf{K}_k=\mathbf{P}_{k \mid k}^{\prime} \mathbf{H}_k^T\left(\mathbf{H}_k \mathbf{P}_{k \mid k}^{\prime} \mathbf{H}_k^T+\mathbf{R}_{\boldsymbol{m}, k}\right)^{-1}
\end{equation}

\subsubsection{Voxel-aware Point Management}
\label{sec:voxel-aware}
This section describes how new map points are registered to update the scene map. Using the estimated camera pose from Sec. \ref{sec:select_map_points}, we triangulate newly extracted feature correspondences into 3D space and integrate them into the voxel map based on their spatial coordinates. During tracking, event noise inevitably creates incorrect feature matches. Adding these erroneous points to the scene map would compromise subsequent state updates. To address this issue, we propose a voxel-aware point management strategy that mitigates event noise through three mechanisms:
1) temporal consistency validation: Due to the typically inconsistent nature of event noise, our system retains only the most frequently tracked points within each voxel, prioritizing stable observations over temporary outliers.
2) spatial proximity culling: To eliminate transient artifacts, we remove points that fall too close to existing map points using a minimum distance threshold, thereby reducing redundancy and enhancing point distinctiveness.
3) voxel capacity regulation: To maintain uniform 3D point distribution, we limit each voxel to five points. Once a voxel reaches this capacity, we automatically exclude any additional points within its boundaries.

\subsubsection{Keyframe Selection and Marginalization}

The system determines keyframe selection by measuring parallax between the current frame and most recent keyframe. When the parallax exceeds 20 pixels, a new keyframe is created while the oldest one is removed from the sliding window. For computational efficiency and real-time operation, frames without sufficient parallax are automatically discarded.


\begin{table}[t]
\setlength\tabcolsep{2pt}
\centering
\renewcommand{\arraystretch}{1.2} 
\caption{Datasets of all sequences for evaluation}
\begin{tabular}{lcccc}
\toprule[1pt]
\textbf{Dataset} & \textbf{Sequence} & \textbf{Duration(min:sec)} & \textbf{Distance(m)} & \textbf{Resolution} \\ 
\midrule
\multirow{5}{*}{VECtor} 
& corner\_slow & 0:41 & 0.87 & 640×480 \\
& board\_slow & 0:34 & 3.10 & 640×480 \\
& desk\_normal & 1:31 & 8.59 & 640×480 \\
& hdr\_normal & 1:00 & 3.11 & 640×480 \\
& mountain\_normal & 1:01 & 8.15 & 640×480 \\
& robot\_normal & 0:40 & 4.33 & 640×480 \\
& sofa\_normal & 1:31 & 30.92 & 640×480 \\
\midrule
\multirow{7}{*}{DSEC} 
& City04\_a & 0:30 & 208.79 & 640×480 \\
& City04\_b & 0:13 & 65.34 & 640×480 \\
& City04\_c & 0:58 & 569.02 & 640×480 \\
& City04\_d & 0:46 & 521.76 & 640×480 \\
& City04\_e & 0:14 & 143.41 & 640×480 \\
& City11\_a & 0:23 & 126.05 & 640×480 \\
& City11\_b & 1:36 & 600.28 & 640×480 \\
\midrule
\multirow{5}{*}{RPG} 
& bin & 0:27 & 8.41 & 240×180 \\
& box & 0:42 & 18.50 & 240×180 \\
& desk & 0:41 & 12.55 & 240×180 \\
& monitor & 0:22 & 6.25 & 240×180 \\
& reader & 0:09 & 4.97 & 240×180 \\
\bottomrule[1pt]
\end{tabular}
\label{tab:sequence_specs}
\end{table}

\section{EXPERIMENTS}
\label{section:experiment}
We evaluated our system using three publicly available datasets that include stereo event cameras and IMUs. We used the Absolute Trajectory Error (ATE) \cite{ATE} as our evaluation metric, computed with the toolkit from \cite{grupp2017evo}. All experiments ran on a computer with an Intel Core i7-12700KF CPU and 32 GB of RAM.

\subsection{Baselines and Datasets}

\subsubsection{Baselines}
We evaluate our Voxel-ESVIO framework against several leading stereo event-based VO systems:

\begin{itemize}
\item \textbf{ESVO}\cite{esvo}: A pioneering event-based stereo VO system using a tracking-and-mapping pipeline.
\item \textbf{ESIO}\cite{esvio}: An indirect method that uses motion compensation to correct event stream distortions and implements sliding-window graph optimization. For a fair comparison, we evaluate only its event-based functionality, without RGB data.
\item \textbf{ESVIO}\cite{esvio2}: A direct approach that enhances pose estimation by incorporating IMU pre-integration constraints in backend optimization.
\item \textbf{ESVIO\_AA}\cite{esvio_aa}: A direct method that uses IMU gyroscope measurements as prior information and introduces an adaptive accumulation event representation method.
\item \textbf{ESVO2}~\cite{esvo2}: An enhanced version of ESVIO\_AA that reduces computational time by optimizing only linear velocity and IMU biases in its streamlined backend.
\end{itemize}


\subsubsection{Datasets}
Table~\ref{tab:sequence_specs} provides a summary of all datasets used in our evaluation, including sequence names, durations, total travel distances, and image resolutions. 
The evaluation utilizes the following datasets:
\begin{itemize}
\item \textbf{VECtor}\cite{vector}: A handheld dataset that captures diverse motion dynamics and illumination conditions across both small and large-scale environments.
\item \textbf{DSEC}\cite{dsec}: An automotive dataset containing various outdoor driving scenarios under different conditions.
\item \textbf{rpg}~\cite{rpg}: A handheld indoor dataset featuring small-scale movements and dynamic scenes.
\end{itemize}

\subsection{Comparison with State-of-the-Art Methods}
As shown in Table~\ref{tab:comparison_val}, we evaluate Voxel-ESVIO against five leading state-of-the-art methods. For ESVIO~\cite{esvio2}, the comparison is limited to available trajectories since it provides raw data for only some sequences and lacks an open-source implementation. Missing data is marked with "–" in the table. We use "failed" to indicate sequences where the open-source  systems failed to run successfully.

Our method achieves the lowest ATE across all sequences, surpassing existing approaches. This superior accuracy results from Voxel-ESVIO's efficient retrieval of map points that combine noise resistance with high observability, enabling more robust geometric constraints.

\begin{table}[tbp]
\setlength\tabcolsep{4pt}
\centering
\renewcommand{\arraystretch}{1.2} 
\caption{RMSE of ATE Compared with State-of-the-Art (Unit: m) }
\begin{tabular}{ccccccc}
\toprule[1pt]
\textbf{Sequence} & \textbf{ESVO} & \textbf{ESIO} & \textbf{ESVIO} & \textbf{ESVIO\_AA} & \textbf{ESVO2} & \textbf{Ours} \\ 
\midrule
corner\_slow & 0.137 & 0.027 & - & 0.055 & \underline{0.022} & \textbf{0.016}  \\ 
board\_slow & failed & 0.230 & - & failed & failed & \textbf{0.021}  \\
desk\_normal & 0.208 & 0.741 & - & 0.191 & \underline{0.165} & \textbf{0.063}  \\ 
hdr\_normal & 0.184 & 0.279 & - & 0.161 & \underline{0.135} & \textbf{0.059}  \\ 
mountain\_normal & 0.389 & 0.265 & - & 0.458 & \underline{0.236} & \textbf{0.021}  \\
robot\_normal & 0.073 & 0.052 & - & 0.152 & \underline{0.048} & \textbf{0.031}  \\ 
sofa\_normal  & failed & 0.439 & - & failed & \underline{0.403} & \textbf{0.155} \\ 
\midrule
City04\_a & 3.70 & 9.40 & 2.01 & 1.04 & \underline{0.48} & \textbf{0.39}  \\ 
City04\_b & 1.15 & 4.34 & \underline{0.48} & 0.67 & 0.80 & \textbf{0.46}  \\ 
City04\_c & 9.32 & 11.53 & 14.01 & 6.37 & \underline{4.66} & \textbf{2.50}  \\ 
City04\_d & 26.76 & 68.22 & - & 7.32 & \underline{5.94} & \textbf{2.36} \\ 
City04\_e & 7.92 & 10.36 & 3.32 & 1.16 & \underline{0.75} & \textbf{0.42}  \\ 
City11\_a & 3.66 & 1.07 & 4.06 & 0.96 & \underline{0.87} & \textbf{0.76}  \\ 
City11\_b & 32.41 & 3.00 & - & 8.70 & \underline{4.42} & \textbf{1.43}  \\ 
\midrule
bin & 0.028 & 0.071 & 0.023 & 0.059 & \underline{0.028} & \textbf{0.020}  \\ 
box & 0.058 & 0.114 & 0.044 & 0.067 & \underline{0.038} & \textbf{0.034}  \\ 
desk  & 0.032 & 0.032 & 0.020 & 0.053 & \underline{0.024} & \textbf{0.020} \\ 
monitor & 0.033 & 0.079 & 0.035 & 0.028 & \underline{0.022} & \textbf{0.019}  \\ 
reader & 0.066 & failed & - & \underline{0.039} & 0.071 & \textbf{0.035}  \\ 
\bottomrule[1pt]
\end{tabular}
\begin{tablenotes} 
\item \textbf{Denotations:} “-” indicates missing experimental results. “failed”  means the open-source system couldn't execute successfully. \textbf{bold} denotes the 1st-ranked method, and \underline{\hspace{0.25cm}} indicates the 2nd-ranked one.
\end{tablenotes}
\label{tab:comparison_val}
\end{table}

\subsection{Time Consumption}
In Table~\ref{tab:time_consumption_detailed}, we evaluate Voxel-ESVIO's runtime performance by measuring the execution time (in milliseconds) of its four main modules: (a) Event Processing, (b) Tracking, (c) MSCKF based Odometry, and (d) Voxel Map Management, along with the total processing time for each stereo event pair. All timing measurements are averaged across the entire sequence. 
Results demonstrate that our method achieves real-time performance with average processing times of 15–25 ms and a maximum runtime of 30 ms across all sequences. Furthermore, experimental results show that the voxel map management module adds minimal computational overhead to the VIO system.

We also compare our total processing time against two representative open-source approaches—ESIO (representing feature-based methods) and ESVO2 (representing direct methods). As shown in Table~\ref{tab:time_study}, our approach achieves significantly lower processing times than both alternatives.

\begin{table}[t]
\setlength\tabcolsep{4pt}
\centering
\renewcommand{\arraystretch}{1.2}
\caption{Detailed time consumption statistics per module (Unit: ms)}
\begin{tabular}{lccccc}
\toprule[1pt]
\textbf{Sequence} & \textbf{(a)} & \textbf{(b)} & \textbf{(c)} & \textbf{(d)} & \textbf{\makecell[c]{Total Time \\ Consumption}} \\
\midrule
corner\_slow & 5.57 & 4.16 & 3.07 & 1.77 & 17.46 \\
board\_slow & 4.83 & 3.53 & 9.27 & 1.75 & 21.04 \\
desk\_normal & 4.89 & 3.17 & 5.05 & 1.89 & 17.35 \\
hdr\_normal & 5.89 & 3.77 & 2.58 & 2.19 & 17.28 \\
mountain\_normal & 5.37 & 3.68 & 5.53 & 2.49 & 19.62 \\
robot\_normal & 4.85 & 3.82 & 4.79 & 1.84 & 17.62 \\
sofa\_normal & 5.16 & 6.31 & 3.26 & 2.04 & 17.77 \\
\midrule
City04\_a & 6.37 & 11.35 & 6.08 & 3.09 & 29.04 \\
City04\_b & 6.49 & 7.06 & 4.10 & 3.62 & 21.73 \\
City04\_c & 5.72 & 11.69 & 3.06 & 2.52 & 25.52 \\
City04\_d & 5.53 & 12.16 & 2.80 & 2.34 & 25.18 \\
City04\_e & 5.87 & 9.72 & 3.32 & 3.36 & 24.57 \\
City11\_a & 5.43 & 10.93 & 6.76 & 3.59 & 28.63 \\
City11\_b & 6.37 & 11.54 & 6.48 & 3.45 & 30.12 \\
\midrule
Bin & 1.44 & 2.74 & 6.83 & 2.32 & 15.17 \\
Boxes & 0.78 & 6.81 & 5.44 & 2.62 & 17.19 \\
Desk & 0.90 & 4.34 & 5.17 & 2.27 & 13.66 \\
Monitor & 0.81 & 4.24 & 6.70 & 4.01 & 16.62 \\
Reader & 0.76 & 4.41 & 7.70 & 4.17 & 17.26 \\
\bottomrule[1pt]
\end{tabular}
\label{tab:time_consumption_detailed}
\begin{tablenotes} 
\item \textbf{Denotations:} (a) Event Processing, (b) Tracking, (c) MSCKF based Odometry, and (d) Voxel Map Management.
\end{tablenotes}
\label{tab:comparison_time}
\end{table}

\begin{table}[t]
\setlength\tabcolsep{12pt}
\centering
\renewcommand{\arraystretch}{1.2}
\caption{Time Consumption Compared with State-of-the-art (Unit: ms)}
\begin{tabular}{lccc}
\toprule[1pt]
\textbf{Sequence} & \textbf{ESIO} & \textbf{ESVO2} & \textbf{Ours} \\
\midrule
corner\_slow & 103.65 & 129.19 & \textbf{17.46} \\
board\_slow & 104.79 & 138.93 & \textbf{21.04} \\
desk\_normal & 104.51 & 195.32 & \textbf{17.35} \\
hdr\_normal & 106.33 & 93.08 & \textbf{17.28} \\
mountain\_normal & 106.73 & 56.12 & \textbf{19.62} \\
robot\_normal & 104.18 & 38.78 & \textbf{17.62} \\
sofa\_normal & 108.20 & 160.56 & \textbf{17.77} \\
\midrule
City04\_a & 77.64 & 43.87 & \textbf{29.04} \\
City04\_b & 75.31 & 115.24 & \textbf{21.73} \\
City04\_c & 77.13 & 91.93 & \textbf{25.52} \\
City04\_d & 77.35 & 65.01 & \textbf{25.18} \\
City04\_e & 77.42 & 80.35 & \textbf{24.57} \\
City11\_a & 73.79 & 107.07 & \textbf{28.63} \\
City11\_b & 76.41 & 82.42 & \textbf{30.12} \\
\midrule
Bin & 99.28 & 18.37 & \textbf{15.17} \\
Boxes & 102.69 & 27.33 & \textbf{17.19} \\
Desk & 101.09 & 25.13 & \textbf{13.66} \\
Monitor & 99.13 & 19.31 & \textbf{16.62} \\
Reader & 105.06 & 29.44 & \textbf{17.26} \\
\bottomrule[1pt]
\end{tabular}
\label{tab:time_study}
\end{table}

\subsection{Ablation Stuty}
To validate our key contributions, we conducted ablation studies across all sequences listed in Table~\ref{tab:sequence_specs}. We compared our full system's performance (labeled as "Ours" in Table~\ref{tab:ablation_study}) against variants with specific features disabled.

\begin{table}[t]
\setlength\tabcolsep{3pt}
\centering
\renewcommand{\arraystretch}{1.15}
\caption{Ablation Study on RMSE of ATE (Unit: m)}
\begin{tabular}{lccc}
\toprule[1pt]
\textbf{Sequence} & \textbf{\makecell[c]{w/o Voxel-based Point \\ Selection}} & \textbf{\makecell[c]{w/o Voxel-aware Point \\ Management}} & \textbf{Ours} \\
\midrule
corner\_slow & 0.046 & 0.031 & \textbf{0.016} \\
board\_slow & 0.059 & 0.047 & \textbf{0.021} \\
desk\_normal & 0.112 & 0.085 & \textbf{0.063} \\
hdr\_normal & 0.067 & \textbf{0.056} & 0.059 \\
mountain\_normal & 0.062 & 0.057 & \textbf{0.021} \\
robot\_normal & 0.035 & 0.043 & \textbf{0.031} \\
sofa\_normal & 0.220 & 0.182 & \textbf{0.155} \\
\midrule
City04\_a & 0.80 & 0.63 & \textbf{0.39} \\
City04\_b & 0.43 & 1.09 & \textbf{0.46} \\
City04\_c & 2.73 & 4.26 & \textbf{2.50} \\
City04\_d & 2.52 & 3.01 & \textbf{2.36} \\
City04\_e & 0.52 & 0.90 & \textbf{0.42} \\
City11\_a & 0.85 & 0.80 & \textbf{0.76} \\
City11\_b & 2.63 & 1.92 & \textbf{1.43} \\
\midrule
bin & 0.025 & 0.020 & \textbf{0.020} \\
box & 0.044 & 0.037 & \textbf{0.034} \\
desk & 0.025 & 0.023 & \textbf{0.020} \\
monitor & 0.020 & 0.021 & \textbf{0.019} \\
reader & 0.039 & 0.036 & \textbf{0.035} \\
\bottomrule[1pt]
\end{tabular}
\label{tab:ablation_study}
\end{table}

\begin{figure*}[t]
\centering
\includegraphics[width=0.96\textwidth]
{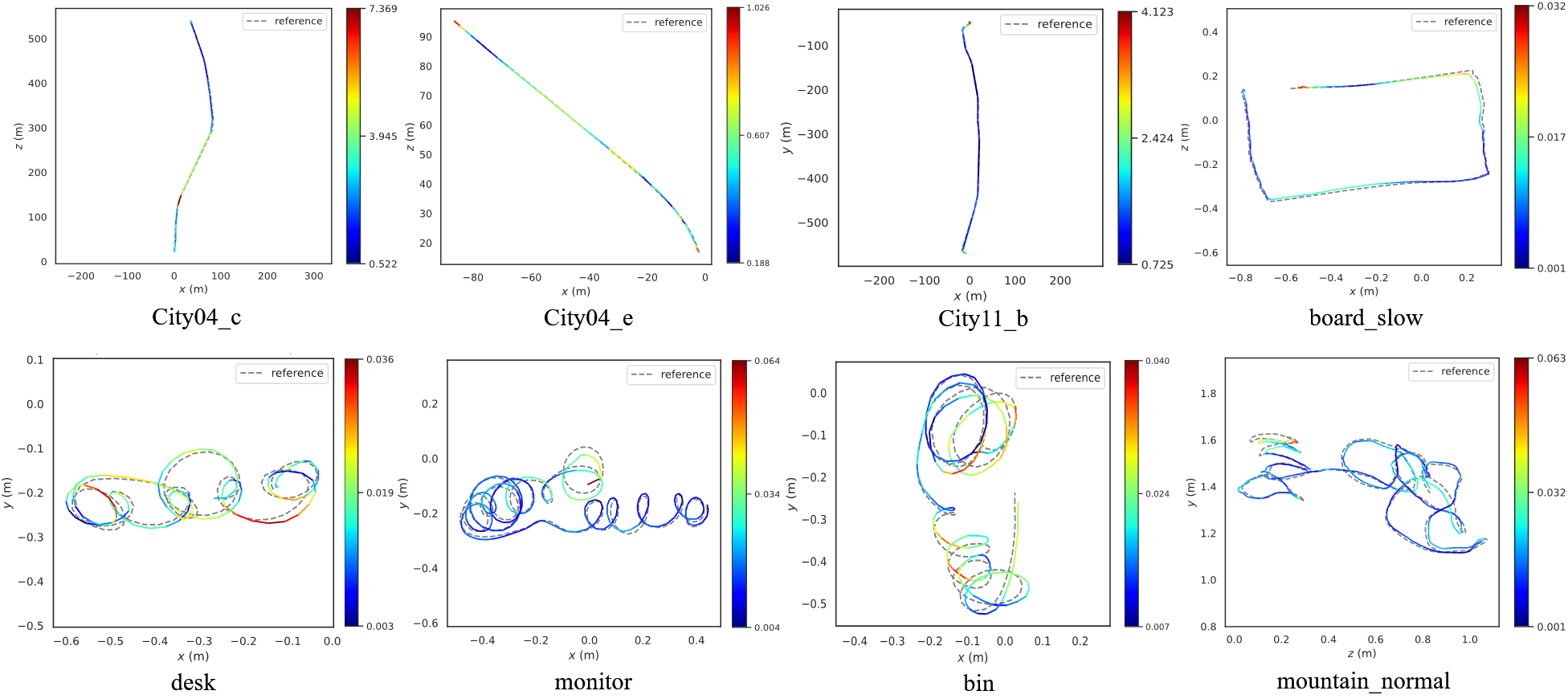} 
\caption{Visualization comparison of estimated trajectory and groundtruth.}
\label{resultcompare}
\end{figure*}

\subsubsection{Effect of Voxel-based Point Selection}

As introduced in Sec. \ref{sec:select_map_points}, voxel-based point selection enables Voxel-ESVIO to efficiently obtain noise-resilient map points with the highest observation likelihood through visited voxels. To evaluate this strategy's effectiveness, we compared ATE results with and without voxel-based point selection. In the baseline configuration (denoted as "w/o voxel-based point selection"), the system selects map points sequentially from storage while maintaining the same total point count as the reference configuration. Results in Table~\ref{tab:ablation_study} demonstrate that voxel-based point selection consistently improves performance.

\subsubsection{Effect of Voxel-aware Point Management}
As detailed in Sec. \ref{sec:voxel-aware}, voxel-aware point management analyzes event noise distribution in both temporal and spatial domains and implements three cascading filters—temporal consistency filtering, spatial proximity culling, and voxel capacity constraint—to eliminate noise-generated points and enhance map point quality.
As shown in Table~\ref{tab:ablation_study}, experimental results demonstrate that integrating the voxel-aware management strategy into the map point registration process yields consistent improvements in estimation accuracy across all test scenarios.

\subsection{Visualization for Trajectory}
Fig. \ref{resultcompare} presents a quantitative evaluation of our trajectory estimates against ground truth measurements across the VECtor, DSEC, and rpg datasets. The results show strong alignment between our estimated paths and reference trajectories, with nearly exact matches in all cases.

\section{CONCLUSIONS}
\label{section:conclusion}
In this work, we propose Voxel-ESVIO, an event-based stereo visual-inertial odometry system that introduces voxel map management to select high-quality map points to mitigate the propagation of event noise. By introducing voxel map management, our system organizes event-generated 3D points in a spatially discretized manner. We developed two key strategies—voxel-based point selection and voxel-aware point management—that work in concert to retrieve noise-resilient map points with the highest observation likelihood. Experiments on public datasets show that our method outperforms existing approaches in both accuracy and computational efficiency. Our results highlight the effectiveness of voxel-level reasoning in enhancing the reliability of event-based visual-inertial odometry.












\bibliographystyle{IEEEtran}
\bibliography{IEEEabrv,mylib}


\end{document}